\documentclass[a4paper,fleqn]{cas-sc}

\usepackage[numbers]{natbib}

\usepackage{algorithm}      
\usepackage{algorithmic}    
\usepackage{subfig}         
\usepackage{float}          

\begin{document}

\let\WriteBookmarks\relax


\shorttitle{QGB-W$k$NN: Quantum Granular-Ball Learning for Robust Classification}

\shortauthors{Yuan et al.}


\title[mode=title]{Reliability-Aware Quantum Granular-Ball Learning for Robust Classification}


\author[1,3,4]{Suzhen Yuan}
\fnmark[1]
\ead{yuansuzhen@cqupt.edu.cn}

\author[2]{Dehang Chen}
\fnmark[1]
\ead{S240201100@stu.cqupt.edu.cn}

\author[3,5]{Lifeng Shen*}
\ead{shenlf@cqupt.edu.cn}

\author[3,5]{Shuyin Xia}
\fnmark[1]
\ead{xiasy@cqupt.edu.cn}

\author[4]{Jeremiah D. Deng}
\fnmark[1]
\ead{jeremiah.deng@otago.ac.nz}


\fntext[1]{These authors contributed equally to this work.}



\affiliation[1]{
    organization={School of Electronic Science and Engineering, Chongqing University of Posts and Telecommunications},
    city={Chongqing},
    postcode={400065},
    country={China}
}

\affiliation[2]{
    organization={School of Computer Science and Technology, Chongqing University of Posts and Telecommunications},
    city={Chongqing},
    postcode={400065},
    country={China}
}

\affiliation[3]{
    organization={School of Artificial Intelligence, Chongqing University of Posts and Telecommunications},
    city={Chongqing},
    postcode={400065},
    country={China}
}

\affiliation[4]{
    organization={School of Computing, University of Otago},
    city={Dunedin},
    postcode={9054},
    country={New Zealand}
}

\affiliation[5]{
    organization={Key Laboratory of Big Data Intelligent Computing, Key Laboratory of Cyberspace Big Data Intelligent Security, Ministry of Education, Chongqing University of Posts and Telecommunications},
    city={Chongqing},
    postcode={400065},
    country={China}
}

\begin{abstract}
Nearest-neighbor classification is widely used in machine learning, yet existing methods often suffer from low computational efficiency and limited robustness in noisy environments. To jointly address these challenges, this paper proposes an efficient and reliable weighted $K$-nearest neighbor classification framework based on quantum granular balls, termed QGB-W$k$NN. The proposed framework improves computational efficiency by integrating quantum-enhanced granular-ball representation with hierarchical nearest-neighbor search, while enhancing classification reliability through a purity-aware weighted decision mechanism. Specifically, quantum-kernel granular balls are constructed to reduce retrieval redundancy and strengthen nonlinear feature representation under limited quantum resources. A granular-ball purity-guided HNSW optimization strategy is developed to exploit structural reliability for hierarchical graph construction during neighbor retrieval, alleviating the local optimality issue caused by conventional random layering. Finally, a weighted voting mechanism jointly incorporating granular-ball similarity and purity is introduced to produce more reliable classification decisions in noisy environments. Extensive experiments on benchmark datasets demonstrate that QGB-W$k$NN achieves competitive classification accuracy while exhibiting favorable Pareto trade-offs between classification performance and computational cost. Moreover, the proposed framework consistently improves robustness under various noisy conditions, suggesting that reliability-aware quantum granular-ball learning provides a promising paradigm for efficient and robust nearest-neighbor classification.
\end{abstract}


\maketitle

\section{Introduction}
The $k$-Nearest Neighbor ($k$-NN) algorithm is a classic and widely applied classification algorithm in machine learning, benefiting from its intuitive working principle, excellent interpretability, and training-free property. It achieves stable and reliable performance in various general classification tasks. However, traditional $k$-NN inherently suffers from two critical defects that limit its practical deployment. First, it requires full traversal of all training samples for similarity calculation during inference, generating substantial redundant computations and leading to low computational efficiency. Second, the standard uniform voting mechanism of $k$-NN is highly susceptible to noise interference, which severely degrades classification performance in noisy scenarios. These two inherent limitations restrict the applicability of conventional $k$-NN in complex real-world tasks \cite{cover1967nearest,bellman2015adaptive,regression1992introduction}.

To solve the computational inefficiency of vanilla $k$-NN, Approximate Nearest Neighbor (ANN) techniques have become a mainstream optimization paradigm. Existing ANN methods are generally divided into three categories, including hashing-based methods, quantization-based methods, and graph-based methods, as systematically reviewed in \cite{azizi2025graph}. Among them, graph-based ANN methods are widely favored due to their superior trade-off between retrieval accuracy and query efficiency. As a representative graph indexing structure, Hierarchical Navigable Small Worlds (HNSW) realizes fast neighbor searching by constructing multi-layered navigable topologies and has become a fundamental benchmark for efficient retrieval tasks \cite{malkov2018efficient}. On this basis, the PECANN framework incorporates HNSW and other graph indexes into clustering tasks, verifying the strong adaptability of graph-based retrieval in complex data scenarios \cite{yu2025pecann}. Furthermore, the SHG method improves search efficiency via hierarchical vector compression and adaptive shortcut connection learning \cite{johnson2019billion,gong2025accelerating}. Although these methods achieve remarkable improvements in retrieval efficiency, they still rely on fine-grained sample-level search. Consequently, the computational overhead remains considerable when handling large-scale datasets.


Existing methods remain confined to sample-level retrieval with prohibitive computational overhead on large-scale data. To overcome this fundamental limitation, the multi-granularity cognitive computing paradigm offers a conceptually distinct solution. Multi-granularity cognitive computing \cite{wang2017dgcc} simulates the human global-first cognitive mechanism, which was pioneered by Wang et al. The core tenet of this theory is to adopt multi-granularity information granules, rather than individual sample points, as the basic computing units, thereby breaking through the intrinsic limitations of conventional point-based computation from a cognitive perspective. As a representative computational model under this paradigm, granular ball computing effectively reduces retrieval redundancy by aggregating discrete raw samples into compact and homogeneous granular units. A variety of optimized granular ball generation strategies have been proposed to boost model learning performance. The GBG++ algorithm enhances the structural stability of granular domains by optimizing generation mechanisms \cite{xie2024gbg++}. The Acc+ method replaces traditional $k$-means clustering with a fast partitioning strategy, effectively improving the construction efficiency of granular balls \cite{xia2022efficient}. The Adp method achieves parameter-free granular ball generation by designing an adaptive purity lower bound. Despite these advances, existing granular-ball methods still suffer from certain limitations. Most methods rely on classical distance metrics and therefore have limited capability in capturing nonlinear correlations in high-dimensional data. Moreover, the robustness of existing granular-ball-based classification methods remains insufficient in noisy environments because their decision processes mainly depend on nearest neighboring granular balls. Overall, traditional $k$-NN, ANN optimization schemes, and existing multi-granularity learning methods are universally constrained by redundant computation, inadequate nonlinear feature extraction, and weak noise resistance.

The integration of quantum computing and machine learning provides a promising research direction to break through these bottlenecks. Owing to the unique advantages of quantum superposition and parallel computing, quantum computing effectively enhances the feature representation and computational efficiency of nearest neighbor algorithms \cite{ciliberto2018quantum}. Current quantum-enhanced $k$-NN research can be divided into two main branches. The first branch focuses on feature mapping enhancement, where diverse quantum distance metrics and quantum kernel methods are designed to mine deep nonlinear data correlations and compensate for the linear representation limitations of traditional models \cite{feng2023enhanced,zardini2024quantum,hoch2025quantum,ronggon2026qknn}. For instance, quantum Hamming distance-based methods have been developed to accelerate nearest neighbor search in high-dimensional binary spaces \cite{li2022quantum}. The second branch concentrates on efficiency optimization, which combines granular ball compression with quantum acceleration to reduce computational complexity \cite{xia2025efficient,yuan2025quantum}. These methods leverage quantum swap tests for nonlinear correlation measurement and utilize quantum parallelism to overcome the traversal computation bottleneck of conventional algorithms \cite{schuld2019quantum,lloyd2022quantum}. However, restricted by limited practical qubit resources, existing quantum granular learning methods cannot simultaneously optimize retrieval efficiency, nonlinear feature extraction, and noise robustness within a unified framework.

Although recent studies have demonstrated that quantum granular-ball representations can effectively improve data compression and retrieval efficiency, reliable decision making remains largely unexplored. Existing quantum nearest-neighbor approaches mainly focus on accelerating similarity computation or reducing search complexity. However, efficient retrieval does not necessarily imply reliable classification. In noisy environments, neighboring samples may contain significant uncertainty, leading to unstable predictions even when retrieval accuracy remains high.

From this perspective, we argue that reliability should be regarded as a first-class component of quantum nearest-neighbor learning. Instead of relying solely on geometric proximity, classification decisions should additionally consider the structural confidence of neighboring regions. Granular-ball purity naturally provides a quantitative measure of such structural reliability. By integrating reliability information into both graph navigation and decision making, classification robustness can be effectively improved.

Motivated by this observation, we propose a reliability-aware quantum granular-ball learning framework, termed QGB-WkNN. Unlike existing quantum granular-ball learning methods that mainly focus on representation construction or retrieval acceleration, QGB-W$k$NN explicitly models reliability throughout the entire learning process, including representation, retrieval, and decision making.
Specifically, the proposed framework consists of three tightly coupled components. First, a quantum-enhanced granular-ball generation module reduces retrieval redundancy while exploiting quantum computing to enhance nonlinear feature representation. Second, a granular-ball purity-guided HNSW optimization strategy exploits granular-ball purity as structural prior knowledge for graph organization and hierarchical optimization. This strategy effectively alleviates the local optimality issue introduced by conventional random layering, thereby further improving nearest-neighbor search efficiency. Finally, a purity-aware weighted classification mechanism jointly considers neighborhood similarity and structural reliability, leading to more robust classification decisions and enhanced robustness in noisy environments. These components establish a unified framework for robust quantum nearest-neighbor classification.
\begin{table}[width=.6\linewidth,cols=2,pos=h]
\centering
\caption{Key Notations Used in the Paper}
\label{tab:notations}
\footnotesize
\begin{tabular*}{\tblwidth}{@{}p{0.7in}L@{}}
\toprule
\textbf{Symbol} & \textbf{Description} \\
\midrule
$k$ & Number of neighbors in $k$-NN \\
$N$ & Total number of samples in dataset \\
$M$ & Dataset size after granular ball compression ($M\ll N$)\\
$GB$ & Granular ball (core data structure) \\
$P$ & Purity of a granular ball (majority-class ratio) \\
$T$ & Purity threshold for granular ball splitting \\
$T_s$ & Similarity threshold for sample assignment \\
$n_q$ & Number of qubits for quantum encoding \\
$\text{sim}_q(\cdot,\cdot)$ & Quantum kernel similarity ($|\langle\psi(x)|\psi(z)\rangle|^2$) \\
$L_{\text{adj}}$ & Adjusted layer number of HNSW node (purity-aware) \\
$\beta$ & Layer adjustment coefficient of HNSW ($0<\beta<1$) \\
$\mathbf{x}_{\text{test}}$ & Test sample to be classified \\
$\rho$ & Gaussian noise ratio in training data \\
\bottomrule
\end{tabular*}
\end{table}


\section{Methodology}
\begin{figure}
    \centering
    \includegraphics[width=0.9\textwidth]{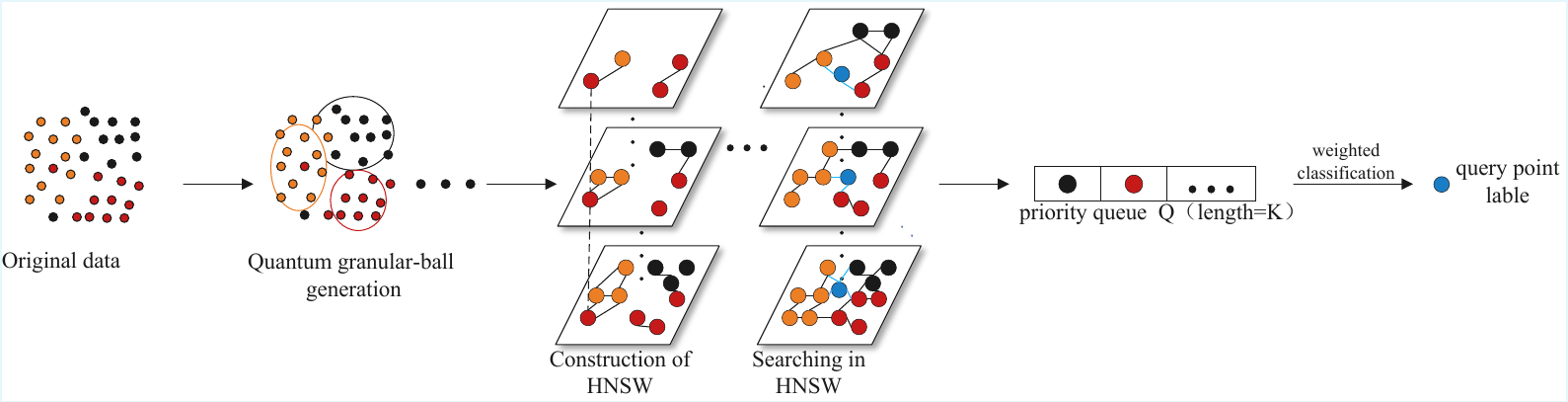}
    \caption{Overall framework figure.}
    \label{fig_overall_framework}
\end{figure}

This section presents the proposed QGB-W$k$NN framework. The goal is to improve approximate $k$NN classification from three aspects: reducing sample-level retrieval redundancy, enhancing nonlinear similarity representation, and improving decision reliability under noisy observations. To avoid notation ambiguity and ensure consistent expression throughout the paper, Table~\ref{tab:notations} summarizes the frequently used symbols. As illustrated in Figure~\ref{fig_overall_framework}, QGB-W$k$NN consists of three coupled components: quantum-kernel granular-ball representation, reliability-guided HNSW graph navigation, and reliability-aware weighted decision learning. The first component compresses raw samples into representative granular balls using quantum-kernel similarity. The second component exploits granular-ball purity to organize the HNSW graph according to structural reliability. The third component fuses granular-ball purity and similarity to perform robust weighted voting.
\subsection{Reliability-Preserving Quantum Granular-ball Representation}
This paper proposes a more efficient method for generating quantum granular balls. By leveraging quantum computing's parallelism, this method improves the efficiency of similarity calculations. Meanwhile, it captures the nonlinear correlations among samples through the inner-product property of quantum states. Algorithm~1 presents the pseudocode of this method. The specific implementation details are elaborated in the following sections.

\begin{algorithm}[t]
    \small
    \caption{Quantum granular ball generation method}
    \label{alg:QMCAGB}
    \textbf{Input}:
    \begin{itemize}
        \item Data set \( D = \{(x_i, y_i)\}, i = 1, \cdots, N \)
        \item Purity threshold \( T \)
        \item Sample assignment similarity threshold \( T_s \)
        \item Quantum circuit parameters \( \theta_{\text{enc}} \) (encoding angles)
    \end{itemize}

    \textbf{Output}: Granular ball set \( GB \)

    \begin{algorithmic}[1]
        \STATE Initialize remaining set \( D_r \leftarrow D \), granular ball set \( GB \leftarrow \emptyset \)
        \WHILE{$D_r \neq \emptyset$}
            \STATE Randomly choose a seed sample from $D_r$:
            $ z \leftarrow D_r[\mathrm{rand}()]$
            \STATE Get \textbf{quantum encoding}: Map \( z \) to \( |\psi_{z}\rangle \), \( \forall s \in D_r \) to \( \{|\psi_s\rangle\} \) via angle encoding
            \STATE Calculate \textbf{quantum similarity}:  \( \text{sim}_q(s, z) \leftarrow |\langle \psi_s | \psi_{z} \rangle|^2,   \forall s \in D_r \)
            \STATE \textbf{Filter} out subset similar to seed $z$: $\mathcal{S}\leftarrow \{s | \mathrm{sim}_q(s,z)\ge T_s, s\in D_r)$
            \STATE Generate \( GB_i = \{z\} \cup \mathcal{S} \)
            \STATE Calculate purity \( P(GB_i) = \frac{\text{majority-label samples in } GB_i}{|GB_i|} \)
            \IF{$P(GB_i) < T$}
                \STATE Split \( GB_i \) into \( K \) class-specific sub-balls \( \{GB_i^1, \dots, GB_i^K\} \)
                \STATE \( GB = GB \cup \{GB_i^1, \dots, GB_i^K\} \)
            \ELSE
                \STATE \( GB = GB \cup \{GB_i\} \)
            \ENDIF
            \STATE \( D_r \leftarrow D_r \setminus GB_i \)
        \ENDWHILE
        \STATE \textbf{return} \( GB \)
    \end{algorithmic}
\end{algorithm}

\subsubsection{Quantum Encoding Preprocessing}
Step 1: Dimensionality reduction. Given the limited qubit resources in quantum computing and the redundancy inherent in high-dimensional data, Linear Discriminant Analysis (LDA) is employed for dimensionality reduction.

\noindent Step 2: Perform normalization. Subsequently, MinMax normalization is implemented to map the data feature values into continuous values within the range $[0, \pi]$, ensuring that each data feature can be directly used as the angle parameter of the quantum rotation gate.

\noindent Step 3: Angle embedding. An angle encoding circuit (i.e., Quantum Feature Mapping, QFM) is constructed to realize the mapping from classical data to quantum states. The quantum feature mapping operator $U_\phi(x)$ is implemented in the form of a tensor product of ${n}_q = \min(d, K-1)$ Pauli-Y rotation gates (Ry gates), i.e., $U_\phi(x) = \bigotimes_{k=0}^{{n}_q-1} R_Y(\theta_k)$. This circuit acts on the initial quantum state $|0\rangle^{\otimes {n}_q}$ (a tensor product of ${n}_q$ ground states $|0\rangle$), converting the classical sample $x$ into the corresponding quantum state $|\psi(x)\rangle = U_\phi(x)|0\rangle^{\otimes {n}_q}$. Here, each Ry gate acts independently on its corresponding qubit, encoding the feature value of the data dimension into the quantum state via a rotation operation.

\subsubsection{Fidelity Calculation}
\begin{figure}
    \centering
    \footnotesize
    \includegraphics[width=0.5\linewidth]{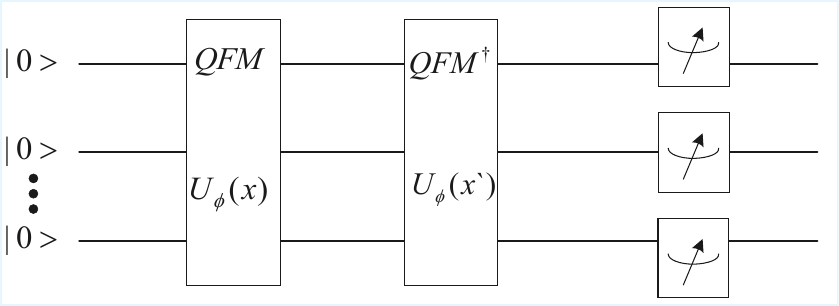}
    \caption{Quantum kernel estimation ansatz.}
    \label{fig_qkm}
\end{figure}

For two classical samples $x$ and $x'$, their encoded quantum states are $|\psi(x)\rangle$ and $|\psi(x')\rangle$, respectively. The quantum kernel estimation ansatz is shown in Fig.~\ref{fig_qkm}, based on which a complete similarity calculation circuit is constructed. The initial state $|0\rangle^{\otimes n_q}$ first passes through $U_\phi(x)$ to generate $|\psi(x)\rangle$, and then through the inverse operator $U_\phi^\dagger(x')$ of the angle encoding circuit for sample $x'$ (where $U_\phi^\dagger(x') = \bigotimes_{k=0}^{n_q-1} R_Y(-\theta'_k)$, and $\theta'_k$ is the normalized angle of the $k$-th dimension feature of $x'$). The probability that the quantum state is $|0\rangle^{\otimes n_q}$ after measurement is recorded, and this probability is exactly the quantum kernel similarity $\text{sim}_q(x, x') = |\langle\psi(x) | \psi(x')\rangle|^2$ between the two samples. This probability ranges from 0 to 1, and the closer the value is to 1, the higher the similarity between the samples.

\subsubsection{Quantum Comparator}
To select candidate samples whose quantum kernel similarity meets the threshold requirement, this paper introduces a quantum comparator. The specific implementation method is as follows.

Quantum kernel similarity is expressed as a quantum state amplitude. It cannot be directly used for comparison operations. Therefore, we first use a Quantum Analog-to-Digital Converter (QADC) to convert the similarity value carried in the amplitude into a $t$-bit binary quantum state encoded by basis states. At the same time, applying the same encoding rule, the similarity threshold $T_s$ is converted into a binary quantum state of the same length. This ensures consistent input formats.

The quantum comparison logic adopts an optimized $t$-bit full quantum comparator. This comparator performs bit-by-bit comparison between the encoded similarity quantum state and the threshold quantum state by cascading quantum logic gates such as CNOT gates and Pauli gates. The comparison order proceeds from the Most Significant Bit (MSB) to the Least Significant Bit (LSB). This circuit has preset clear judgment rules. When the similarity is greater than or equal to the threshold, the output qubit of the comparator is set to the $|1\rangle$ state. When the similarity is less than the threshold, the output qubit is set to the $|0\rangle$ state.

A measurement operation is performed on the output qubit of the comparator. The measurement result is a classical binary value (1 or 0). When the result is 1, it indicates that the similarity between the corresponding candidate sample and the reference sample meets the threshold. When the result is 0, it indicates that the sample does not meet the threshold condition. By counting and statistically analyzing all measurement results, all eligible candidate samples can be selected.

\subsubsection{Iterative Update}
After screening the candidate samples that meet the similarity requirements, the purity of the sample set is calculated. It is then compared with the preset purity threshold. If the purity meets the threshold, a granular ball is generated directly. If the purity does not meet the threshold, the sample set is split by category to generate sub-granular balls. Meanwhile, the center and radius of the granular ball are calculated. The core attributes of the granular ball are recorded. These attributes include the center point, purity, category, and radius.

After generating the current granular ball, the samples contained in it are removed from the remaining sample set. New granular balls are then generated iteratively. When all samples have been assigned to granular balls and no samples remain, the iterative loop terminates. The remaining samples are merged to generate the final granular ball. The entire quantum granular ball generation process is thus completed.

\subsection{Reliability-Guided Graph Navigation}
\begin{figure}
    \centering
    \footnotesize
    \subfloat[]{\includegraphics[scale=0.3]{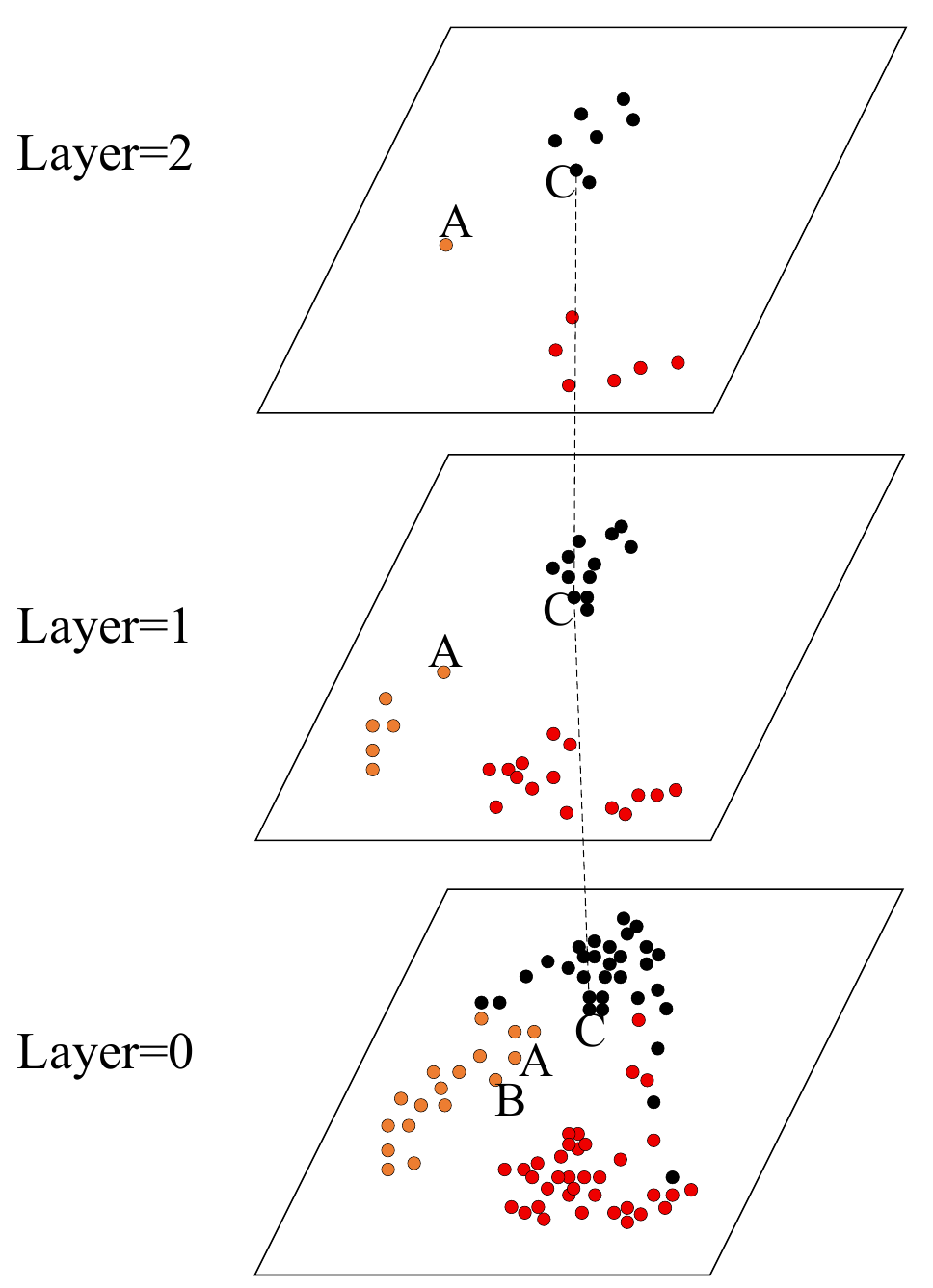}}
    \hspace{8pt}
    \subfloat[]{\includegraphics[scale=0.3]{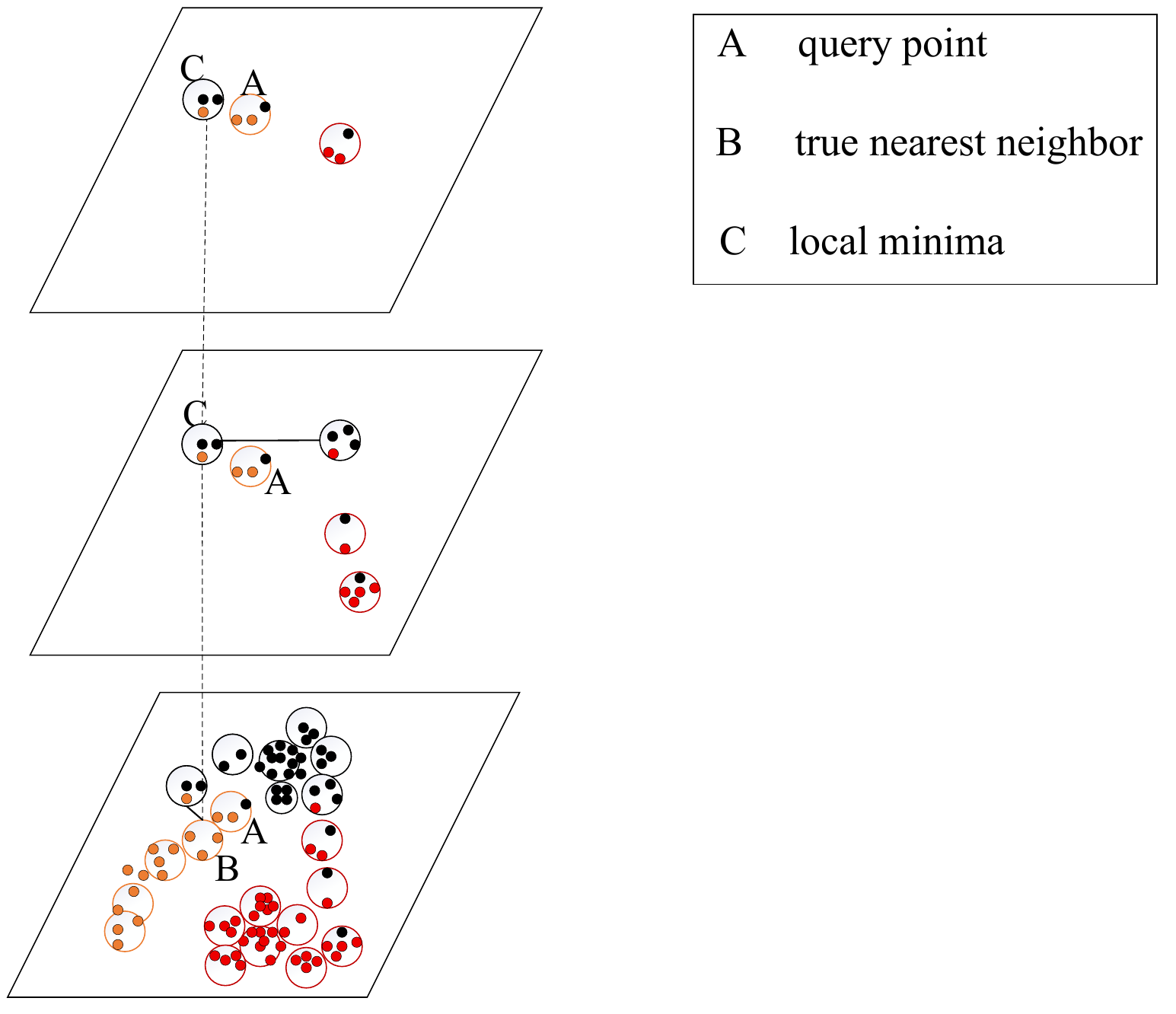}}
    \caption{Illustration of the local optimum problem of Hierarchical Navigable Small World (HNSW) and its mitigation strategy based on granular balls. (a) The local optimum problem of HNSW, where query point A locates the locally optimal nearest neighbor point C. (b) The granular ball-based hierarchical assignment strategy alleviates the local optimum problem of HNSW, enabling query point A to find its true nearest neighbor point B.}
    \label{fig:hnsw_granular_ball_strategy}
\end{figure}

The traditional Hierarchical Navigable Small Worlds (HNSW) algorithm's level assignment strategy relies on random probability to assign levels to nodes. It does not account for data distribution characteristics. This makes the hierarchical structure constructed by the algorithm unable to adapt to differences in data distribution across different regions. Nodes in locally high-density regions have redundant level assignments. Nodes in low-density regions lack sufficient cross-region connections. These problems not only reduce retrieval efficiency but also lead to local optimality. Figure~\ref{fig:hnsw_granular_ball_strategy}(a) visually presents the local optimality problem caused by random level assignment in the traditional HNSW algorithm. To address this flaw, this paper proposes a dynamic level assignment strategy based on granular ball purity. It uses granular-ball purity as an indicator of structural reliability. A high-purity granular ball usually corresponds to a homogeneous and reliable local region, whereas a low-purity granular ball often lies near class boundaries or ambiguous regions. Therefore, incorporating purity into HNSW layer assignment allows the graph hierarchy to reflect the reliability of different granular regions. Figure~\ref{fig:hnsw_granular_ball_strategy}(b) visually verifies the mitigation effect of this granular ball purity-based level assignment strategy on the local optimality phenomenon. Granular ball purity itself is a core indicator in the granular ball splitting process. Using it as the basis for density measurement requires no additional computational overhead. It also avoids the $O(n^2)$ complexity brought by distance matrices in traditional density estimation methods. Thus, it achieves adaptive optimization of the hierarchical structure while ensuring efficiency.

The specific hierarchical logic steps are as follows.

\paragraph{Step 1: Density Aware Layer Calculation.} First, generate an initial layer number via a random function:
\begin{equation}
    L_{\text{temp}} = \left\lfloor -\log_2(r) \right\rfloor
    \label{eq:hnsw_temp_layer}
\end{equation}
where $r$ is a random value in $[0,1]$. Then $L_{\text{temp}}$ is dynamically adjusted based on the purity $P$ of the granular ball to which the node belongs, with a non-negative constraint:
\begin{equation}
    L_{\text{adj}} = \max\left(L_{\text{temp}} - \beta  ( P - P_{\text{avg}}),0 \right)
    \label{eq:hnsw_adj_layer}
\end{equation}
where $P_{\text{avg}}$ is the average purity of all granular balls, and $\beta$ is the adjustment coefficient $(0 < \beta < 1)$.

For high-purity nodes ($P > P_{\text{avg}}$), the number of layers will be appropriately reduced to increase their local connections in lower layers and strengthen the regional connectivity of homogeneous data. For low-purity nodes ($P < P_{\text{avg}}$), the number of layers will be correspondingly increased, enabling them to quickly cross sparse regions through long-distance connections in higher layers and avoid ineffective searches.

\paragraph{Step 2: Layer number boundary constraints.} The final number of layers is determined by $L=\min(L_{\text{adj}},\log M)$ to ensure that it does not exceed the preset maximum number of layers.


\subsection{Reliability-Aware Decision Learning}
Existing $k$NN-based classification methods either rely on a single-granularity ball decision or on equal-weight voting, both of which lack robustness in noisy scenarios. To address this limitation, this paper proposes a weighted classification method. For a given test sample $\mathbf{x}_{\text{test}}$, its predicted class $\hat{y}$ is determined by the weighted voting mechanism:
\begin{equation}
    \hat{y} = \arg\max_{c \in C} \sum_{i \in \mathcal{N}_k(\mathbf{x}_{\text{test}})} w_i \cdot \mathbf{1}\left(c = y_i\right)
    \label{eq:knn_prediction}
\end{equation}
where $\mathcal{N}_k(\mathbf{x}_{\text{test}})$ denotes the $k$ nearest neighbor granular balls, $w_i$ is the comprehensive weight of the $i$-th granular ball, and $\mathbf{1}(\cdot)$ is the indicator function.

The comprehensive weight $w_i$ is jointly determined by granular ball purity, normalized similarity, and weight coefficient:
\begin{equation}
    w_i = P_i~  \text{sim}_i^{(\text{norm})} \alpha
    \label{eq:weight_calc}
\end{equation}

Granular ball purity $P_i$ is used to measure the class consistency of samples within a granular ball. A higher purity value indicates stronger classification reliability of the granular ball. The normalized similarity $\text{sim}_i^{(\text{norm})}$ represents the similarity between the test sample and the center of the granular ball. Its calculation steps are described below. First, calculate the raw similarity:
\begin{equation}
    \text{sim}_i = \frac{1}{1 + \|\mathbf{x}_{\text{test}} - \mathbf{c}_i\|_2 + \epsilon}
    \label{eq:similarity_calc}
\end{equation}
Here $\epsilon$ is an extremely small positive value. It prevents the denominator from being zero. This paper sets the value of $\epsilon=10^{-8}$. Then perform normalization on the similarity values. This step eliminates scale differences across different test samples:
\begin{equation}
    \text{sim}_i^{(\text{norm})} = \frac{\text{sim}_i - \min_{j \in \mathcal{N}_k(\mathbf{x}_{\text{test}})} (\text{sim}_j)}{\max_{j \in \mathcal{N}_k(\mathbf{x}_{\text{test}})} (\text{sim}_j) - \min_{j \in \mathcal{N}_k(\mathbf{x}_{\text{test}})} (\text{sim}_j)}
    \label{eq:similarity_norm}
\end{equation}

All values of $\text{sim}_i^{(\text{norm})}$ are uniformly set to 1 when all similarity values are equal. The weight coefficient $\alpha$ is an adjustable hyperparameter. It is designed to control the amplification effect of purity on the final weight. It is set to the empirical value of 2.2.

After calculating the comprehensive weight $w_i$ of all granular balls, weighted voting is implemented according to the following logic. Accumulate the weights of all granular balls in the $k$-nearest neighbour granular ball set $\mathcal{N}_k(\mathbf{x}_{\text{test}})$ according to their dominant classes. Select the class with the highest total weighted score as the final class of the test sample $\mathbf{x}_{\text{test}}$. Choose the class corresponding to the granular ball with the highest purity when multiple classes obtain the same highest total weighted score. This ensures the reliability of the classification decision.

\subsection{Theoretical Analysis}
The following complexity analysis describes the algorithmic structure under the quantum-kernel estimation model. Since the current experiments are conducted by classical simulation, the empirical runtime does not represent actual quantum hardware acceleration.

\subsubsection{Construction Phase}
The construction phase includes quantum encoding preprocessing, quantum granular ball iterative generation and HNSW construction.

\paragraph{Quantum Encoding Preprocessing.} This step sequentially performs Linear Discriminant Analysis (LDA) dimensionality reduction, linear normalization, and quantum state encoding.

Given \(N\) samples of dimension \(d\), LDA projects them into an \(n_q\)-dimensional space (where \(n_q \ll d\)), with a time complexity of:
\begin{equation}
T_{\text{LDA}} = O(N \cdot d \cdot n_q)
\label{eq:T_lda}
\end{equation}

The time complexity for both linear normalization and quantum state encoding scales linearly with the number of samples and features:
\begin{equation}
T_{\text{norm}} = O(N \cdot n_q), \quad T_{\text{enc}} = O(N \cdot n_q)
\label{eq:T_norm_enc}
\end{equation}

Therefore, the total preprocessing time is:
\begin{equation}
T_{\text{pre}} = O(N \cdot d \cdot n_q) + O(N \cdot n_q) = O(N \cdot d \cdot n_q)
\label{eq:T_pre}
\end{equation}

\paragraph{Quantum Granular Ball Iterative Generation.} Let \(n_i\) denote the number of remaining samples in the \(i\)-th iteration. The time required to calculate the quantum similarities between the benchmark sample and all others is:
\begin{equation}
T_{\text{sim}}^{(i)} = O(n_i \cdot n_q)
\label{eq:T_sim_iter}
\end{equation}

Traversing all \(n_i\) similarity values and comparing them with the similarity threshold costs \(O(n_i)\). Consequently, the time complexity for a single iteration is:
\begin{equation}
T_{\text{iter}}^{(i)} = O(n_i \cdot n_q) + O(n_i) = O(n_i \cdot n_q)
\label{eq:T_single_iter}
\end{equation}

Summing over all \(k_0\) iterations, and noting that \(\sum_{i=1}^{k_0} n_i = N\), the total time becomes:
\begin{equation}
T_{\text{GB}} = \sum_{i=1}^{k_0} O(n_i \cdot n_q) = O\!\left(n_q \sum_{i=1}^{k_0} n_i\right) = O(N \cdot n_q)
\label{eq:T_gb}
\end{equation}

The operation of splitting low-purity granular balls (where the total local sample size satisfies \(\sum n_s \leq N\)) does not alter the asymptotic order of this complexity.

\paragraph{HNSW Construction.} The hierarchical insertion of \(M\) granular ball centers involves a constant number of candidate nodes \(M_h\) per layer, resulting in a time complexity of:
\begin{equation}
T_{\text{insert}} = O(M \cdot n_q \cdot M_h) = O(M \cdot n_q)
\label{eq:T_insert}
\end{equation}

The hierarchical traversal overhead is \(O(M \log M)\). Thus, the total time for HNSW construction is:
\begin{equation}
T_{\text{HNSW}} = O(M \cdot n_q) + O(M \log M)
\label{eq:T_hnsw}
\end{equation}

\paragraph{Overall Construction Complexity.} Integrating Equations~\eqref{eq:T_pre}, \eqref{eq:T_gb}, and \eqref{eq:T_hnsw}, the overall construction time complexity is:
\begin{equation}
T_{\text{construct}} = O(N \cdot d \cdot n_q) + O(N \cdot n_q) + O(M \cdot n_q) + O(M \log M)
\label{eq:T_combine}
\end{equation}

Given that \(n_q \ll d\) and typically \(M \ll N\), the dominant term is:
\begin{equation}
T_{\text{construct}} = O(N \cdot d)
\label{eq:T_final}
\end{equation}

\subsubsection{Search Phase}
The search process of QGB-W$k$NN comprises data processing on the query sample, HNSW-based nearest neighbor search, and granular ball weighted voting.

Given a single query sample of dimension \(d\), data preprocessing applies the same operations as during construction: LDA projection, and normalization. The time complexity for processing one sample is:
\begin{equation}
T_{\text{q\_pre}} = O(d \cdot n_q) + O(n_q) = O(d \cdot n_q)
\label{eq:T_q_pre}
\end{equation}
The dominant term \(O(d \cdot n_q)\) stems from projecting the \(d\)-dimensional query into the \(n_q\)-dimensional subspace via LDA.

The HNSW index is then searched to retrieve the \(k\) nearest granular ball centers to the encoded query. The HNSW search complexity is:
\begin{equation}
T_{\text{hnsw\_search}} = O(n_q \cdot \log M)
\label{eq:T_hnsw_search}
\end{equation}
where the \(n_q\) factor accounts for distance calculations in the \(n_q\)-dimensional quantum feature space at each traversal step.

Finally, for the \(k\) retrieved granular balls, a weighted voting prediction is made based on each ball's purity and similarity to the query. This step requires:
\begin{equation}
T_{\text{vote}} = O(k \cdot n_q)
\label{eq:T_vote}
\end{equation}

The total search time complexity is therefore:
\begin{equation}
\begin{split}
T_{\text{search}} &= O(d \cdot n_q) + O(n_q \cdot \log M) + O(k \cdot n_q) \\
&= O\bigl(n_q \cdot (d + \log M + k)\bigr)
\label{eq:T_search_combine}
\end{split}
\end{equation}
Since \(k\) is a small constant and typically \(M \ll N\), the expression simplifies to:
\begin{equation}
T_{\text{search}} = O(d + \log M)
\label{eq:T_search_final}
\end{equation}

Table~\ref{tab:notations1} summarizes the explanations of the symbols used in Table~\ref{tab:time_complexity}.

\begin{table}[width=0.6\linewidth,cols=3,pos=h]
\centering
\caption{Time Complexity Comparison of Different Methods}
\label{tab:time_complexity}
\small
\setlength{\tabcolsep}{1pt}
\begin{tabular*}{\tblwidth}{LCC}
\toprule[1.2pt]
\textbf{Method} & \textbf{Construction} & \textbf{Search} \\
\midrule[0.8pt]
QGB-W$k$NN  & $O(N \cdot d)$ & $O(d + \log M)$ \\
GB$k$NN++             & $O\left((T + R) \cdot N \cdot d\right)$ & $O(k_g \cdot d)$ \\
HNSW              & $O(N \cdot \log N) $ & $O(d \cdot \log N)$ \\
QSVC              & ${O(N^3)} $ & $O(d\cdot n_q)$ \\
GLVQ              & $O(N\cdot n_p\cdot d)$  & $O(n_p\cdot d)$ \\
\bottomrule[1.2pt]
\end{tabular*}
\end{table}

\begin{table}[width=.5\linewidth,cols=2,pos=h]
\centering
\caption{Explanations of Core Symbols in Time Complexity Comparison}
\label{tab:notations1}
\small
\setlength{\tabcolsep}{1pt}
\begin{tabular*}{\tblwidth}{@{}p{0.7in}L@{}}
\toprule
\textbf{Symbol} & \textbf{Description} \\
\midrule
$k_g$ & Number of subgraphs in DiskANN \\
$n_q$& Number of qubits for angle encoding\\
${n_p}$ & Number of prototype vectors of GLVQ\\
\bottomrule
\end{tabular*}
\end{table}

\section{Experiments}

\subsection{Experimental Setup}

\subsubsection{Data Preparation}
To verify the performance of the proposed method, we selected eight publicly available real-world classification datasets covering multiple domains, including signal detection, text classification, medical diagnosis, and financial evaluation, as well as two synthetic datasets.

Data preprocessing pipeline:
\begin{itemize}
    \item Missing values are imputed with column means to avoid introducing bias.
    \item Categorical features are converted into numerical vectors via OneHotEncoder, with unknown categories ignored.
\end{itemize}

The key information of all datasets used in the experiments is summarized in Table~\ref{tab:dataset_details}.

\begin{table}[width=0.7\linewidth,cols=5,pos=h]
\centering
\caption{Key Information of Experimental Datasets.}
\label{tab:dataset_details}
\footnotesize
\setlength{\tabcolsep}{6pt}
\begin{tabular*}{\tblwidth}{@{}LCCCC@{}}
\toprule[1.5pt]
& \textbf{Dataset} & \textbf{Samples} & \textbf{Dim} & \textbf{Classes} \\
\midrule[1pt]
& Ionosphere          & 351       & 34    & 2 \\
& Spambase            & 4601      & 57    & 2 \\
& German Credit       & 1000      & 20    & 2 \\
real & Diabetes (Pima)     & 768       & 8     & 2 \\
& Breast Cancer       & 569       & 30    & 2 \\
& Mushrooms           & 8124      & 117   & 2 \\
& Splice              & 3190      & 216   & 3 \\
& 20 Newsgroups       & 4532      & 2000  & 5 \\
synthetic & a1       & 1000000   & 30    & 3 \\
& a2       & 1000000   & 40    & 4 \\
\bottomrule[1.5pt]
\end{tabular*}
\end{table}

\subsubsection{Compared Algorithms}
To comprehensively evaluate the classification performance of the proposed method, five representative classification algorithms are selected for comparison. These methods cover conventional nearest-neighbor classification, granular-ball-based classification, graph-based nearest-neighbor search, prototype-based learning, and quantum machine learning. For fair comparison, all methods perform classification on the same training and testing sets. The conventional abbreviations of the compared algorithms are adopted throughout the subsequent experiments.

$k$NN~\citep{cover1967nearest} is the classical $k$-nearest neighbor classification algorithm and serves as the fundamental baseline. It performs classification according to the labels of neighboring samples and is widely used in various machine learning tasks.

GB$k$NN++~\citep{xie2024gbg++} is a representative granular-ball-based classification method. It performs classification on compressed granular-ball representations and exhibits superior efficiency and robustness compared with traditional $k$NN.

GLVQ~\citep{sato1995generalized} is a prototype-based classification algorithm that learns representative prototypes for each class. It has been widely applied to evaluate the effectiveness of feature representation and classification capability.

HNSW~\citep{malkov2018efficient} is a state-of-the-art graph-based nearest-neighbor search method. In this study, HNSW is combined with the standard $k$NN voting strategy to accomplish classification tasks. It provides a strong efficiency-oriented baseline for nearest-neighbor classification.

QSVC~\citep{chowdhury2025quantum} is a representative quantum machine learning classifier based on quantum kernel evaluation. It is adopted to assess the effectiveness of the proposed method relative to existing quantum-enhanced classification approaches.

\subsubsection{Parameter Settings}
The hyperparameters are configured with a purity threshold of 0.8, an initial similarity of 0.3. For the Hierarchical Navigable Small World (HNSW) algorithm, the parameters include the L2 norm as the space type, 8-nearest neighbours for graph construction, and a weight coefficient $\alpha$ of 2.2. All experiments were implemented in Python~3.11 on the PyCharm platform. In the quantum granular ball generation component, the quantum circuits involved, including data encoding and quantum kernel fidelity computation, were executed using the default quantum simulator in the PennyLane~0.43.1 framework, while the remaining quantum computational operations were simulated on a classical computer. The experimental hardware environment comprised a 13th Gen Intel\textregistered\ Core\texttrademark\ i7-13700 processor. Each experiment was repeated three times with a fixed random seed of 42, and the average results were used for performance evaluation.

This paper emphasizes that the quantum operations are simulated classically. Consequently, the computational efficiency reported in the experiments does not include actual quantum speedup; the claimed quantum parallelism remains a theoretical advantage pending future hardware implementation.

\subsection{Results}

\subsubsection{Effectiveness}
Table~\ref{tab:noise_0_percent} reports the classification accuracy and Macro-F1 scores of all compared methods. QGB-W$k$NN achieves the best or highly competitive performance on several real-world datasets, especially on high-dimensional and multi-class tasks such as 20 Newsgroups and Splice. On some relatively simple or highly separable datasets, such as Mushrooms, classical baselines can already reach near-perfect performance, leaving limited room for improvement. Therefore, the advantage of QGB-W$k$NN should not be interpreted as uniformly higher accuracy on every dataset, but as a better overall balance between accuracy, efficiency, and robustness. The results indicate that quantum-kernel granular-ball representation and purity-similarity weighted voting are particularly useful when the data distribution is complex or affected by noise.

\begin{table}[width=\textwidth,cols=13,pos=h]
\centering
\caption{Comparison on classification accuracy and macro-F1. Bold indicates the best result. ``---'' denotes memory overflow or timeout.}
\label{tab:noise_0_percent}
\setlength{\heavyrulewidth}{1.2pt}
\resizebox{\tblwidth}{!}{
\begin{tabular}{lcccccccccccc}
\toprule
Dataset & \multicolumn{2}{c}{QGB-W$k$NN} & \multicolumn{2}{c}{GB$k$NN} & \multicolumn{2}{c}{GLVQ} & \multicolumn{2}{c}{HNSW} & \multicolumn{2}{c}{$k$NN} & \multicolumn{2}{c}{QSVC} \\
& Acc & Macro-F1 & Acc & Macro-F1 & Acc & Macro-F1 & Acc & Macro-F1 & Acc & Macro-F1 & Acc & Macro-F1 \\
\midrule
20newsgroups & \textbf{0.919$\pm$0.006} & \textbf{0.918} & 0.538$\pm$0.026 & 0.539 & 0.859$\pm$0.005 & 0.862 & 0.866$\pm$0.008 & 0.868 & 0.866$\pm$0.008 & 0.868 & --- & --- \\
breastcancer & \textbf{0.962$\pm$0.010} & \textbf{0.959} & 0.933$\pm$0.036 & 0.928 & \textbf{0.962$\pm$0.020} & \textbf{0.959} & 0.953$\pm$0.022 & 0.949 & 0.953$\pm$0.022 & 0.949 & 0.848$\pm$0.045 & 0.830 \\
diabetes & \textbf{0.766$\pm$0.030} & \textbf{0.751} & 0.678$\pm$0.043 & 0.655 & 0.749$\pm$0.029 & 0.708 & 0.747$\pm$0.007 & 0.715 & 0.747$\pm$0.007 & 0.715 & 0.716$\pm$0.008 & 0.627 \\
german\_credit & 0.717$\pm$0.040 & \textbf{0.691} & 0.618$\pm$0.050 & 0.570 & \textbf{0.722$\pm$0.028} & 0.608 & 0.717$\pm$0.008 & 0.615 & 0.717$\pm$0.008 & 0.615 & --- & --- \\
ionosphere & \textbf{0.873$\pm$0.037} & \textbf{0.860} & 0.826$\pm$0.035 & 0.783 & 0.812$\pm$0.064 & 0.762 & 0.822$\pm$0.080 & 0.774 & 0.822$\pm$0.080 & 0.774 & 0.808$\pm$0.064 & 0.763 \\
mushrooms & 0.953$\pm$0.038 & 0.951 & 0.938$\pm$0.011 & 0.938 & \textbf{1.000$\pm$0.000} & \textbf{1.000} & \textbf{1.000$\pm$0.000} & \textbf{1.000} & \textbf{1.000$\pm$0.000} & \textbf{1.000} & --- & --- \\
spambase & 0.905$\pm$0.006 & 0.900 & 0.856$\pm$0.021 & 0.844 & \textbf{0.923$\pm$0.006} & \textbf{0.919} & 0.918$\pm$0.008 & 0.913 & 0.918$\pm$0.008 & 0.913 & --- & --- \\
splice & \textbf{0.898$\pm$0.060} & \textbf{0.887} & 0.536$\pm$0.007 & 0.539 & 0.861$\pm$0.005 & 0.861 & 0.806$\pm$0.008 & 0.807 & 0.806$\pm$0.008 & 0.807 & --- & --- \\
easy\_3class\_30dim & 0.993$\pm$0.000 & 0.993 & \textbf{0.997$\pm$0.000} & \textbf{0.997} & \textbf{0.997$\pm$0.000} & \textbf{0.997} & \textbf{0.997$\pm$0.000} & \textbf{0.997} & \textbf{0.997$\pm$0.000} & \textbf{0.997} & --- & --- \\
ultra\_easy\_40dim & \textbf{1.000$\pm$0.000} & \textbf{1.000} & \textbf{1.000$\pm$0.000} & \textbf{1.000} & \textbf{1.000$\pm$0.000} & \textbf{1.000} & \textbf{1.000$\pm$0.000} & \textbf{1.000} & \textbf{1.000$\pm$0.000} & \textbf{1.000} & --- & --- \\
\bottomrule
\end{tabular}
}
\end{table}
\subsubsection{Pareto Trade-off}
\begin{figure}
    \centering
    \includegraphics[width=.9\linewidth]{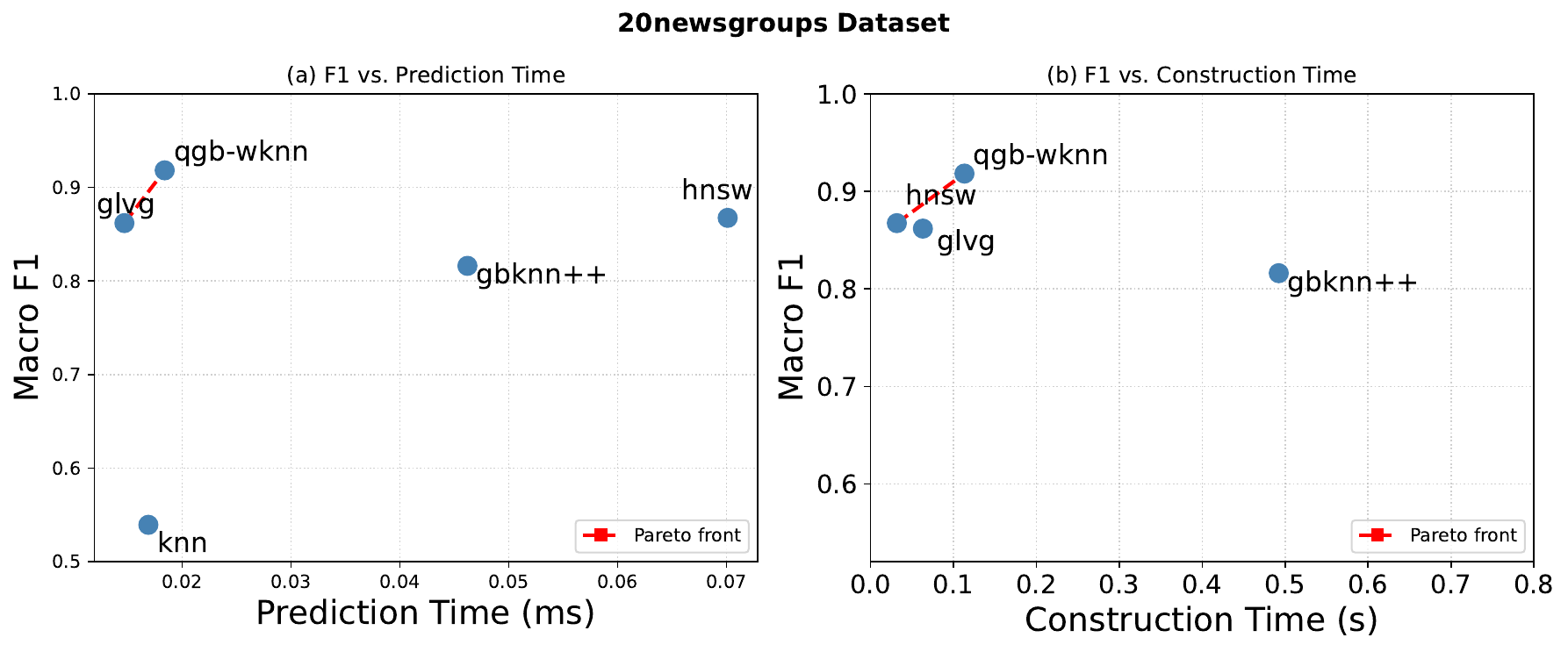}
    \caption{Pareto trade-off between classification accuracy and computational overhead of different methods on the 20newsgroups dataset during model construction and inference stages. Some baseline methods are not presented in the Pareto diagrams because their excessive computational consumption or inferior classification performance leads to completely dominated solutions, which provides no valid comparison significance. The proposed QGB-W$k$NN achieves a better balance between accuracy and efficiency.}
    \label{fig_pareto_20newsgroups}
\end{figure}

\begin{figure}
    \centering
    \includegraphics[width=.9\linewidth]{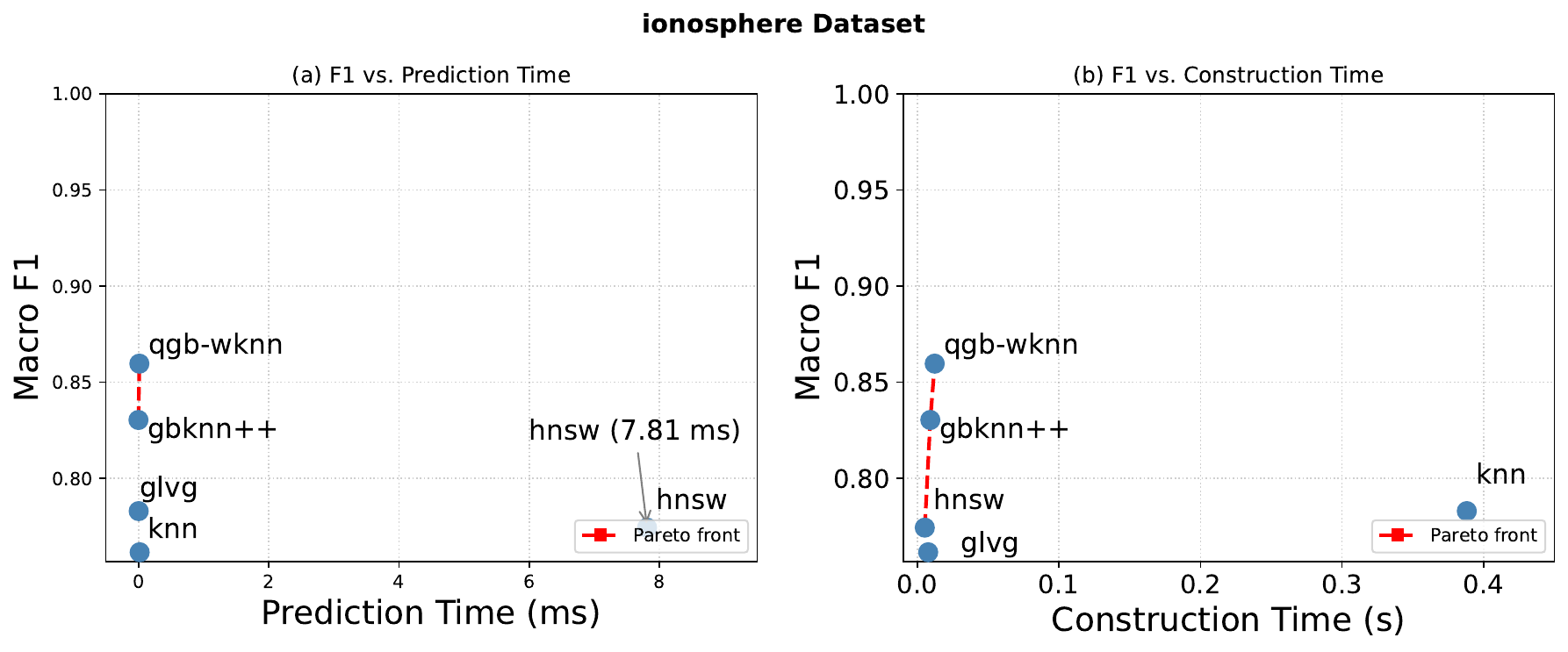}
    \caption{Pareto trade-off between classification accuracy and computational overhead of different methods on the Ionosphere dataset during model construction and inference stages. Several baseline methods are omitted here since their performance is fully dominated by other competitors, resulting in no effective Pareto optimal solutions for comparison. QGB-W$k$NN maintains a favorable balance between classification performance and computational cost.}
    \label{fig_pareto_ionosphere}
\end{figure}

To further evaluate the comprehensive performance of all methods, we select two representative datasets: 20newsgroups and Ionosphere. They are used to analyze the Pareto trade-off between classification performance and computational consumption. Figure~\ref{fig_pareto_20newsgroups} and Figure~\ref{fig_pareto_ionosphere} present the results on these two datasets, respectively. The figures intuitively show the balance between classification accuracy and computational overhead during model construction and inference. Traditional baseline methods have inherent limitations in Pareto optimization. They cannot effectively balance accuracy and efficiency. These methods either sacrifice accuracy to reduce computational cost, or maintain high precision at a high computational expense. In comparison, QGB-W$k$NN achieves a favorable trade-off on both datasets, demonstrating better comprehensive adaptability than the baselines.

\subsubsection{Robustness}
As shown in Tables~\ref{tab:noise_20_percent} and~\ref{tab:noise_40_percent}, QGB-W$k$NN exhibits stable performance under different levels of Gaussian noise. Although some classical baselines may obtain the best accuracy on highly separable datasets, QGB-W$k$NN maintains competitive accuracy and Macro-F1 across different noise ratios. This indicates that the proposed reliability-aware voting mechanism can reduce the influence of noisy or ambiguous granular balls and improve the overall robustness of nearest-neighbor classification.

\begin{table}[width=\textwidth,cols=13,pos=h]
\centering
\caption{Classification accuracy and macro-F1 under noise ratio $\boldsymbol{0.2}$. Bold indicates the best result. ``---'' denotes memory overflow or timeout.}
\label{tab:noise_20_percent}
\setlength{\heavyrulewidth}{1pt}
\resizebox{\tblwidth}{!}{%
\begin{tabular}{lcccccccccccc}
\toprule
Dataset & \multicolumn{2}{c}{QGB-W$k$NN} & \multicolumn{2}{c}{GB$k$NN} & \multicolumn{2}{c}{GLVQ} & \multicolumn{2}{c}{HNSW} & \multicolumn{2}{c}{$k$NN} & \multicolumn{2}{c}{QSVC} \\
& Acc & Macro-F1 & Acc & Macro-F1 & Acc & Macro-F1 & Acc & Macro-F1 & Acc & Macro-F1 & Acc & Macro-F1 \\
\midrule
20newsgroups & \textbf{0.917$\pm$0.005} & \textbf{0.916} & 0.823$\pm$0.005 & 0.824 & 0.855$\pm$0.003 & 0.858 & 0.861$\pm$0.007 & 0.863 & 0.861$\pm$0.007 & 0.863 & --- & --- \\
breastcancer & \textbf{0.965$\pm$0.000} & \textbf{0.962} & 0.930$\pm$0.015 & 0.924 & 0.962$\pm$0.013 & 0.959 & 0.962$\pm$0.020 & 0.959 & 0.962$\pm$0.020 & 0.959 & 0.789$\pm$0.035 & 0.771 \\
diabetes & \textbf{0.764$\pm$0.032} & \textbf{0.749} & 0.723$\pm$0.027 & 0.684 & 0.755$\pm$0.031 & 0.715 & 0.745$\pm$0.016 & 0.713 & 0.745$\pm$0.016 & 0.713 & --- & --- \\
german\_credit & 0.725$\pm$0.043 & \textbf{0.701} & 0.670$\pm$0.005 & 0.593 & 0.722$\pm$0.014 & 0.584 & \textbf{0.727$\pm$0.033} & 0.626 & \textbf{0.727$\pm$0.033} & 0.626 & --- & --- \\
ionosphere & \textbf{0.873$\pm$0.028} & \textbf{0.861} & 0.864$\pm$0.008 & 0.841 & 0.803$\pm$0.070 & 0.746 & 0.822$\pm$0.080 & 0.774 & 0.822$\pm$0.080 & 0.774 & 0.761$\pm$0.014 & 0.721 \\
mushrooms & 0.958$\pm$0.048 & 0.957 & \textbf{1.000$\pm$0.000} & \textbf{1.000} & 0.9998$\pm$0.0004 & 0.9998 & \textbf{1.000$\pm$0.000} & \textbf{1.000} & \textbf{1.000$\pm$0.000} & \textbf{1.000} & --- & --- \\
spambase & 0.900$\pm$0.007 & 0.894 & 0.866$\pm$0.003 & 0.862 & \textbf{0.908$\pm$0.010} & \textbf{0.902} & 0.903$\pm$0.006 & 0.898 & 0.903$\pm$0.006 & 0.898 & --- & --- \\
splice & 0.867$\pm$0.028 & 0.858 & \textbf{0.911$\pm$0.004} & \textbf{0.907} & 0.844$\pm$0.017 & 0.844 & 0.809$\pm$0.012 & 0.810 & 0.809$\pm$0.012 & 0.810 & --- & --- \\
Average & \textbf{0.871$\pm$0.081} & \textbf{0.872} & 0.848$\pm$0.092 & 0.830 & 0.856$\pm$0.085 & 0.826 & 0.854$\pm$0.084 & 0.830 & 0.854$\pm$0.084 & 0.830 & --- & --- \\
\bottomrule
\end{tabular}
}

\end{table}
\begin{table}[width=\textwidth,cols=13,pos=h]
\centering
\caption{Classification accuracy and macro-F1 under noise ratio $\boldsymbol{0.4}$. Bold indicates the best result. ``---'' denotes memory overflow or timeout.}
\label{tab:noise_40_percent}
\setlength{\heavyrulewidth}{1pt}
\resizebox{\tblwidth}{!}{%
\begin{tabular}{lcccccccccccc}
\toprule
Dataset & \multicolumn{2}{c}{QGB-W$k$NN} & \multicolumn{2}{c}{GB$k$NN} & \multicolumn{2}{c}{GLVQ} & \multicolumn{2}{c}{HNSW} & \multicolumn{2}{c}{$k$NN} & \multicolumn{2}{c}{QSVC} \\
& Acc & Macro-F1 & Acc & Macro-F1 & Acc & Macro-F1 & Acc & Macro-F1 & Acc & Macro-F1 & Acc & Macro-F1 \\
\midrule
20newsgroups & \textbf{0.895$\pm$0.017} & \textbf{0.894} & 0.807$\pm$0.010 & 0.808 & 0.845$\pm$0.001 & 0.848 & 0.850$\pm$0.002 & 0.851 & 0.850$\pm$0.002 & 0.851 & --- & --- \\
breastcancer & 0.950$\pm$0.005 & 0.947 & 0.947$\pm$0.026 & 0.943 & 0.953$\pm$0.010 & 0.949 & \textbf{0.956$\pm$0.018} & \textbf{0.953} & \textbf{0.956$\pm$0.018} & \textbf{0.953} & 0.681$\pm$0.022 & 0.662 \\
diabetes & \textbf{0.742$\pm$0.053} & \textbf{0.731} & 0.684$\pm$0.023 & 0.652 & 0.740$\pm$0.028 & 0.702 & 0.714$\pm$0.017 & 0.674 & 0.714$\pm$0.017 & 0.674 & --- & --- \\
german\_credit & 0.712$\pm$0.042 & \textbf{0.681} & 0.673$\pm$0.037 & 0.610 & \textbf{0.743$\pm$0.032} & 0.608 & 0.693$\pm$0.024 & 0.582 & 0.693$\pm$0.024 & 0.582 & --- & --- \\
ionosphere & 0.850$\pm$0.008 & 0.838 & \textbf{0.864$\pm$0.022} & \textbf{0.845} & 0.808$\pm$0.064 & 0.757 & 0.831$\pm$0.061 & 0.790 & 0.831$\pm$0.061 & 0.790 & 0.723$\pm$0.045 & 0.695 \\
mushrooms & 0.986$\pm$0.008 & 0.985 & 0.989$\pm$0.003 & 0.989 & 0.996$\pm$0.002 & 0.996 & \textbf{1.000$\pm$0.000} & \textbf{1.000} & \textbf{1.000$\pm$0.000} & \textbf{1.000} & --- & --- \\
spambase & 0.867$\pm$0.009 & 0.860 & 0.841$\pm$0.005 & 0.835 & 0.890$\pm$0.007 & 0.883 & \textbf{0.893$\pm$0.002} & \textbf{0.887} & \textbf{0.893$\pm$0.002} & \textbf{0.887} & --- & --- \\
splice & \textbf{0.871$\pm$0.035} & \textbf{0.865} & 0.834$\pm$0.013 & 0.824 & 0.811$\pm$0.026 & 0.806 & 0.781$\pm$0.020 & 0.780 & 0.781$\pm$0.020 & 0.780 & --- & --- \\
Average & \textbf{0.859$\pm$0.083} & \textbf{0.850} & 0.830$\pm$0.097 & 0.814 & 0.848$\pm$0.084 & 0.819 & 0.840$\pm$0.087 & 0.815 & 0.840$\pm$0.087 & 0.815 & --- & --- \\
\bottomrule
\end{tabular}
}
\end{table}
\subsubsection{Parameter Sensitivity Analysis}
\begin{figure}
    \centering
    \footnotesize
    \subfloat[]{\includegraphics[scale=0.3]{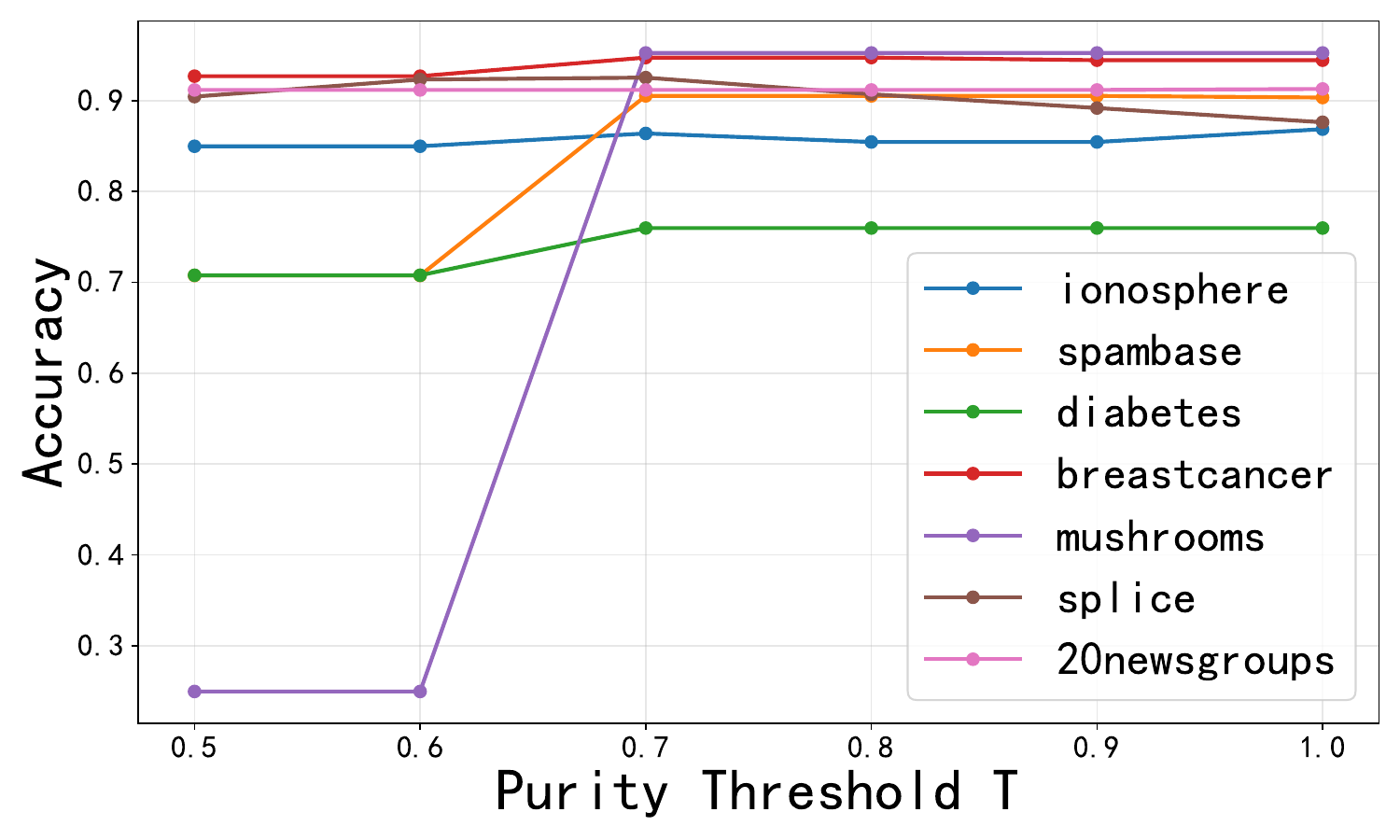}}
    \hspace{8pt}
    \subfloat[]{\includegraphics[scale=0.3]{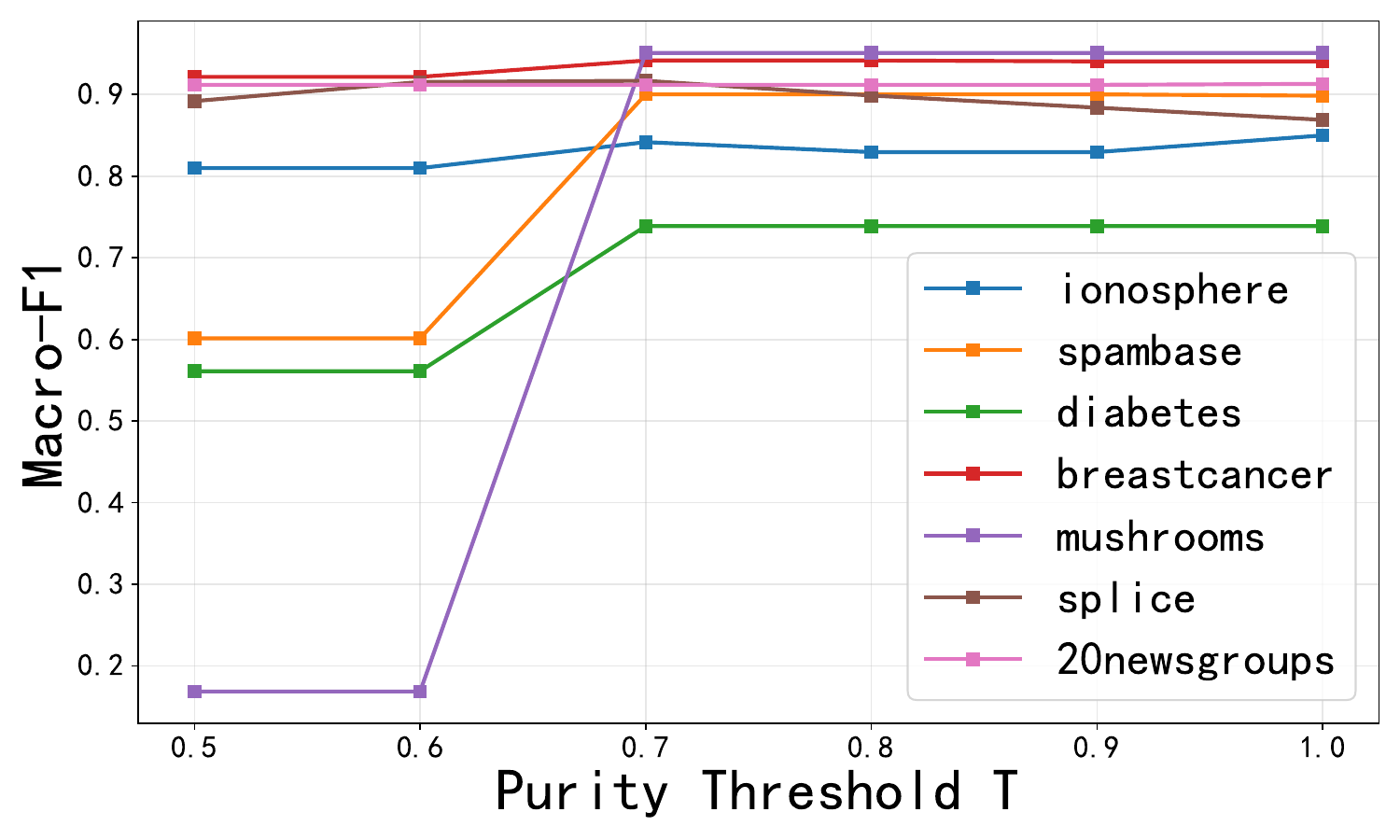}}
    \\
    \vspace{10pt} 
    \subfloat[]{\includegraphics[scale=0.3]{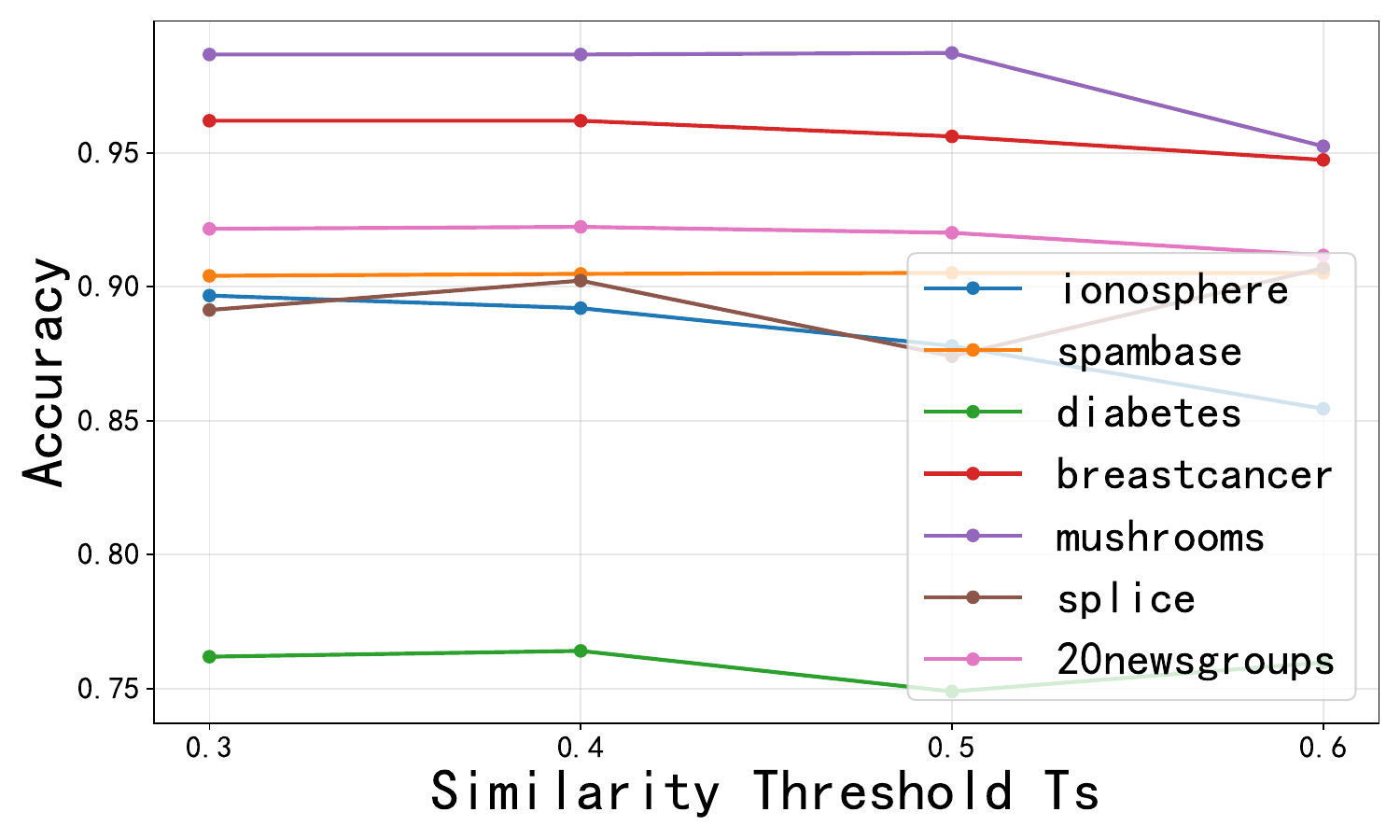}}
    \hspace{8pt}
    \subfloat[]{\includegraphics[scale=0.3]{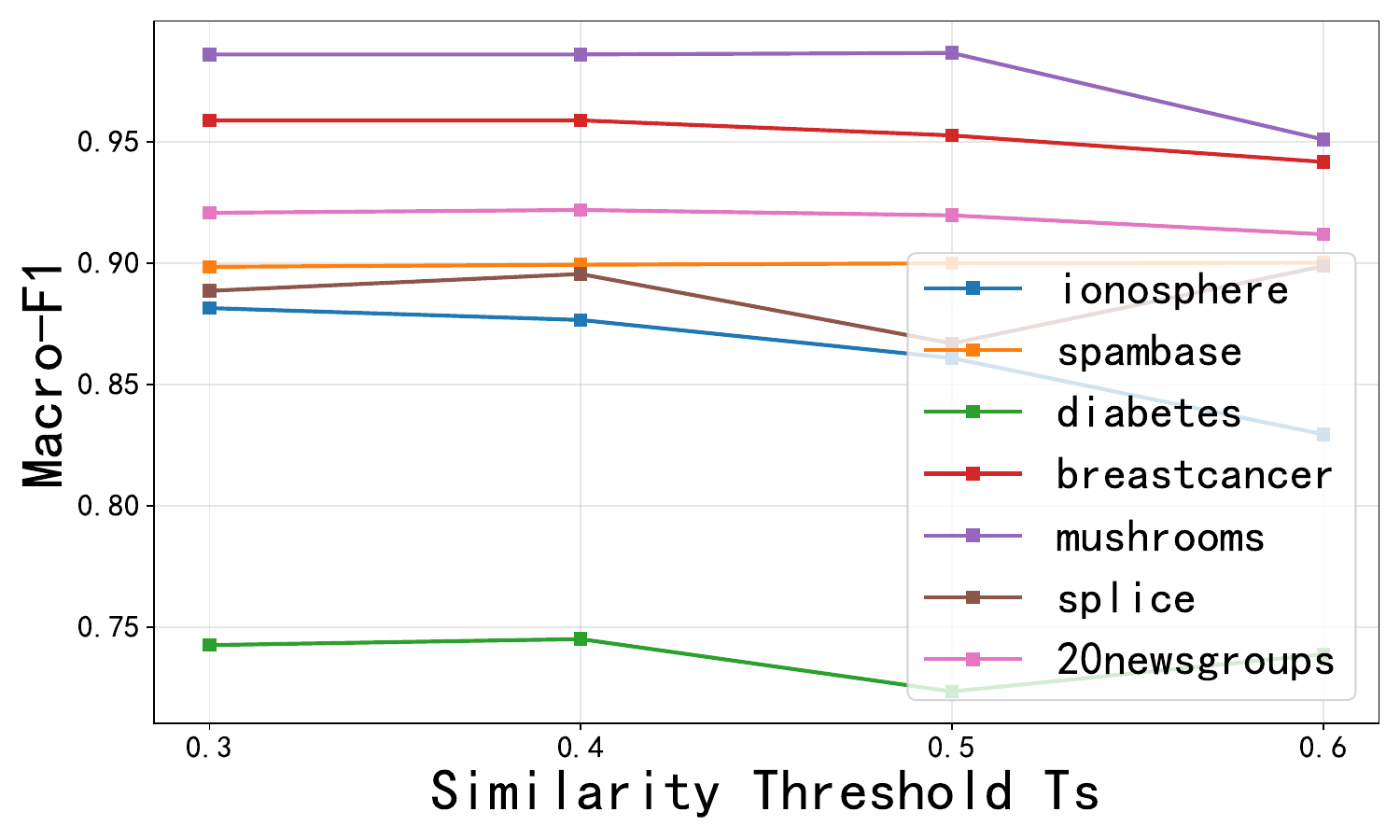}}
    \caption{Parameter sensitivity analysis of the proposed QGB-W$k$NN. 
    (a) Classification accuracy with respect to the purity threshold $T$.
    (b) Macro-F1 score with respect to the purity threshold $T$.
    (c) Classification accuracy with respect to the similarity threshold $T_s$.
    (d) Macro-F1 score with respect to the similarity threshold $T_s$.}
    \label{fig:parameter_sensitivity}
\end{figure}

We perform parameter sensitivity analysis on two key hyperparameters of the proposed method: the granular ball purity threshold $T$ and the similarity threshold $T_s$. A series of real datasets are used to test the variation of classification accuracy and Macro-F1 score, and the results are presented in Figure~\ref{fig:parameter_sensitivity}. The purity threshold $T$ is assigned with values from 0.5 to 1.0, including 0.5, 0.6, 0.7, 0.8, 0.9 and 1.0, and the similarity threshold $T_s$ is set to 0.3, 0.4, 0.5 and 0.6. It can be observed that the model is sensitive to $T$ when $T < 0.8$, and its performance increases steadily along with the rising threshold. Once $T \ge 0.8$, the performance tends to be stable and is no longer obviously affected by minor changes of $T$. Within the valid range of $T_s$, the model performance gradually degrades as $T_s$ increases, and the best overall performance is obtained at $T_s=0.3$. Based on the comprehensive performance and parameter robustness across all datasets, we finally set $T=0.8$ and $T_s=0.3$ in our experiments.

The above trends can be reasonably explained from the perspective of granular ball characteristics. Increasing the purity threshold helps construct high-quality granular balls, which can better describe data distribution and distinguish different categories. When $T$ exceeds 0.7, the quality of granular balls has reached a near-optimal state, so further raising the threshold brings limited gains. In terms of the similarity threshold $T_s$, an excessively large threshold will filter out too many valid samples during granular ball construction, resulting in insufficient information and degraded representation ability. Selecting $T_s=0.3$ balances sample retention and screening criteria, which is conducive to stable and high classification performance.

\subsection{Ablation Study}
An ablation study is carried out to evaluate the contributions of the purity-guided HNSW layer assignment and the weighted classification mechanism in QGB-W$k$NN to classification accuracy and noise robustness, involving the following ablation variants:
\begin{itemize}
\item \textbf{Basic}: the standard HNSW method without any modifications.
\item \textbf{P-$k$NN}: a variant of QGB-W$k$NN in which the weighted classification mechanism is removed.
\item \textbf{W-$k$NN}: a variant of QGB-W$k$NN in which the purity-guided HNSW layer assignment module is removed.
\end{itemize}

We tested all variants on multiple datasets under different noise ratios, and the detailed experimental results are presented in Table~\ref{tab:four_methods} and Table~\ref{tab:macro_recall}. The ablation results show that the different proposed modules contribute from different perspectives. The weighted voting mechanism mainly improves classification robustness by reducing the influence of low-purity or weakly related granular balls. The purity-guided HNSW layer assignment mainly affects the organization of the search graph and contributes to the efficiency-robustness trade-off. It is worth noting that a single module may obtain slightly higher accuracy on certain datasets, because different datasets have different distributions and noise patterns. However, the full QGB-W$k$NN framework provides a more stable overall balance across accuracy, Macro-F1, search efficiency, and noise robustness, which is consistent with the design goal of robust large-scale nearest-neighbor classification.

\begin{table}[H,width=0.8\linewidth,cols=6,pos=h]
\centering
\caption{Accuracy of Basic, P-$k$NN, W-$k$NN and QGB-W$k$NN under various noise ratios on datasets.}
\label{tab:four_methods}
\small
\setlength{\tabcolsep}{1pt}
\begin{tabular*}{\tblwidth}{LCRRRR}
\toprule[1.5pt]
Dataset & Noise Ratio & Basic & P-$k$NN & W-$k$NN & QGB-W$k$NN \\
\midrule[1pt]

\multirow{3}{*}{Ionosphere}
& 0     & 0.822 & 0.859 & \textbf{0.887} & \textbf{0.873} \\
& 0.2   & 0.822 & 0.831 & 0.873 & \textbf{0.873} \\
& 0.4   & 0.831 & 0.831 & 0.873 & \textbf{0.850} \\

\multirow{3}{*}{Spambase}
& 0     & 0.918 & 0.919 & 0.912 & 0.905 \\
& 0.2   & 0.903 & 0.908 & 0.897 & 0.900 \\
& 0.4   & \textbf{0.893} & 0.858 & 0.872 & 0.867 \\

\multirow{3}{*}{German\_credit}
& 0     & 0.717 & 0.730 & 0.745 & \textbf{0.717} \\
& 0.2   & \textbf{0.727} & 0.695 & 0.715 & 0.725 \\
& 0.4   & 0.693 & 0.715 & \textbf{0.775} & 0.712 \\

\multirow{3}{*}{Diabetes}
& 0     & 0.747 & 0.766 & 0.779 & \textbf{0.766} \\
& 0.2   & 0.745 & 0.792 & 0.792 & \textbf{0.764} \\
& 0.4   & 0.714 & 0.766 & 0.786 & \textbf{0.742} \\

\multirow{3}{*}{Breastcancer}
& 0     & 0.953 & 0.974 & 0.974 & \textbf{0.962} \\
& 0.2   & 0.962 & 0.956 & 0.965 & \textbf{0.965} \\
& 0.4   & \textbf{0.956} & 0.945 & 0.956 & 0.950 \\

\multirow{3}{*}{Mushrooms}
& 0     & \textbf{1.000} & 0.999 & 0.989 & 0.953 \\
& 0.2   & \textbf{1.000} & 1.000 & 0.999 & 0.958 \\
& 0.4   & \textbf{1.000} & 1.000 & 0.996 & 0.986 \\

\multirow{3}{*}{Splice}
& 0     & 0.806 & 0.815 & 0.934 & \textbf{0.898} \\
& 0.2   & 0.809 & 0.781 & \textbf{0.940} & 0.867 \\
& 0.4   & 0.781 & 0.734 & 0.919 & \textbf{0.871} \\

\multirow{3}{*}{20newsgroups}
& 0     & 0.866 & 0.865 & 0.900 & \textbf{0.919} \\
& 0.2   & 0.861 & 0.862 & 0.931 & \textbf{0.917} \\
& 0.4   & 0.850 & 0.836 & 0.906 & \textbf{0.895} \\
\bottomrule[1.5pt]
\end{tabular*}
\end{table}

\begin{table}[width=0.8\linewidth,cols=6,pos=h]
\centering
\caption{Macro-recall of Basic, P-$k$NN, W-$k$NN and QGB-W$k$NN under various noise ratios on datasets.}
\label{tab:macro_recall}
\small
\setlength{\tabcolsep}{1pt}
\begin{tabular*}{\tblwidth}{LCRRRR}
\toprule[1.5pt]
Dataset & Noise Ratio & Basic & P-$k$NN & W-$k$NN & QGB-W$k$NN \\
\midrule[1pt]

\multirow{3}{*}{Ionosphere}
& 0     & 0.774 & 0.800 & 0.849 & \textbf{0.860} \\
& 0.2   & 0.774 & 0.769 & 0.847 & \textbf{0.861} \\
& 0.4   & 0.790 & 0.769 & 0.820 & \textbf{0.838} \\

\multirow{3}{*}{Spambase}
& 0     & \textbf{0.913} & 0.903 & 0.900 & 0.900 \\
& 0.2   & 0.898 & 0.892 & 0.900 & \textbf{0.894} \\
& 0.4   & \textbf{0.887} & 0.835 & 0.850 & 0.860 \\

\multirow{3}{*}{German\_credit}
& 0     & 0.615 & 0.617 & 0.644 & \textbf{0.691} \\
& 0.2   & 0.626 & 0.582 & 0.658 & \textbf{0.701} \\
& 0.4   & 0.582 & 0.611 & 0.686 & \textbf{0.681} \\

\multirow{3}{*}{Diabetes}
& 0     & 0.715 & 0.726 & 0.741 & \textbf{0.751} \\
& 0.2   & 0.713 & 0.763 & \textbf{0.780} & 0.749 \\
& 0.4   & 0.674 & 0.735 & \textbf{0.769} & 0.731 \\

\multirow{3}{*}{Breastcancer}
& 0     & 0.949 & 0.970 & 0.957 & \textbf{0.959} \\
& 0.2   & 0.959 & 0.960 & \textbf{0.962} & \textbf{0.962} \\
& 0.4   & \textbf{0.953} & \textbf{0.962} & 0.955 & 0.947 \\

\multirow{3}{*}{Mushrooms}
& 0     & \textbf{1.000} & \textbf{1.000} & 0.981 & 0.951 \\
& 0.2   & \textbf{1.000} & \textbf{1.000} & 0.999 & 0.957 \\
& 0.4   & \textbf{1.000} & \textbf{1.000} & 0.996 & 0.985 \\

\multirow{3}{*}{Splice}
& 0     & 0.807 & 0.771 & 0.919 & \textbf{0.887} \\
& 0.2   & 0.810 & 0.738 & 0.920 & \textbf{0.858} \\
& 0.4   & 0.780 & 0.674 & 0.930 & \textbf{0.865} \\

\multirow{3}{*}{20newsgroups}
& 0     & 0.868 & 0.865 & 0.910 & \textbf{0.918} \\
& 0.2   & 0.863 & 0.861 & 0.920 & \textbf{0.916} \\
& 0.4   & 0.851 & 0.839 & 0.910 & \textbf{0.894} \\
\bottomrule[1.5pt]
\end{tabular*}
\end{table}

\section{Conclusion and Limitations}

In this paper, we proposed QGB-W$k$NN, a reliability-aware quantum granular-ball framework for robust nearest-neighbor classification. The proposed method reformulates conventional sample-level $k$NN classification as a quantum-enhanced, structure-aware granule-level learning paradigm. By introducing granular-ball purity into HNSW layer assignment, the search graph becomes aware of the structural reliability of different data regions. By combining granular-ball similarity with  purity in weighted voting, the final decision process becomes less sensitive to noisy or ambiguous neighboring granules.

Theoretical analysis shows that, under the quantum-kernel estimation model and with $M\ll N$ granular balls, QGB-W$k$NN can reduce the search burden from sample-level retrieval to granule-level graph search. Experiments on real-world and synthetic benchmark datasets demonstrate that the proposed method achieves competitive classification accuracy, improved robustness under Gaussian noise, and enhanced trade-offs between classification performance and computational cost. The ablation results further show that the purity-guided graph navigation and reliability-aware voting mechanisms contribute to robustness and the overall efficiency--accuracy balance from different perspectives.

The present study still has certain limitations. The current experiments are conducted using classical simulation of quantum circuits, and therefore the reported results cannot reflect the runtime performance on real quantum hardware. Future work will further validate the proposed method on practical quantum devices under realistic noise conditions.

\section*{Acknowledgements}
The authors would like to acknowledge the financial support of the National Key R\&D Program of China (2025YFF0514702, 2025YFF0514700),  the Chongqing Natural Science Foundation Innovation and Development Joint Fund (CSTB2025NSCQ-LZX0141), and the project "Research and Development of Quantum Technology-Based Cultural Relics Detection Technology."
\section*{CRediT authorship contribution statement}
\textbf{Suzhen Yuan}: Writing – review \& editing, Formal analysis, Conceptualization. \textbf{Dehang Chen}: Writing – original draft, Writing – review \& editing, Visualization, Validation, Methodology, Data curation. \textbf{Lifeng Shen}: Writing – review \& editing, Supervision. \textbf{Shuyin Xia}: Supervision. \textbf{Jeremiah D. Deng}: Supervision, Writing – review \& editing. 

\section*{Declaration of competing interest}
The authors declare that they have no known competing financial interests or personal relationships that could have appeared to 
influence the work reported in this paper.

\section*{Data availability}
Data will be made available on request.
\printcredits

\bibliographystyle{unsrtnat} 

\bibliography{cas-refs}
\end{document}